%% file: main.tex
\pdfoutput=1

\documentclass[11pt]{article}

\usepackage[]{EMNLP2023}

\usepackage{times}
\usepackage{latexsym}
\usepackage{soul}

\usepackage[T1]{fontenc}

\usepackage[utf8]{inputenc}

\usepackage{microtype}

\usepackage{inconsolata}

\usepackage{xspace}

\usepackage{color}
\usepackage{amsfonts}
\usepackage{amsmath}

\usepackage{algorithm}
\usepackage{algorithmic}

\usepackage{graphicx} 
\usepackage{float} 
\usepackage{subfigure}

\usepackage{multirow}
\usepackage{booktabs}

\usepackage{rotating}

\newcommand{\ours}{\textsc{PiLLow}\xspace}

\title{\ours: Enhancing Efficient Instruction Fine-tuning \\ via Prompt Matching}

\author{Zhenting Qi\thanks{~~Equal Contributions.}~~$^{\clubsuit,\heartsuit}$ \hspace{0.5em} Xiaoyu Tan$^{*~\heartsuit}$\thanks{~~Corresponding author.} \hspace{0.5em} Shaojie Shi$^{\diamondsuit}$ \hspace{0.5em} Chao Qu$^{\heartsuit}$ \hspace{0.5em} Yinghui Xu$^\spadesuit$ \hspace{0.5em} Yuan Qi$^\spadesuit$ \\
$^\clubsuit$ University of Illinois at Urbana-Champaign \quad $^\heartsuit$ INF Technology (Shanghai) Co., Ltd. \quad \\ $^\spadesuit$ AI$^3$ Institute, Fudan University \quad $^\diamondsuit$ Shanghai University of Engineering Science \\
\texttt{zhenting.19@intl.zju.edu.cn, yulin.txy@inftech.ai}
}

\begin{document}

\maketitle

\begin{abstract}

Instruction fine-tuning has conventionally been employed to adapt Large Language Models (LLMs) to a variety of tasks. Nonetheless, this technique often necessitates substantial computational resources, making it impractical for deployment by individuals or small-scale entities. Recently, Low-Rank Adaptation (LoRA) has become a promising alternative, offering high capabilities on par with full tuning with reduced resource overhead. However, attaining satisfactory performance through the fine-tuning of LoRA is a non-trivial challenge. In this paper, we propose \ours, which aims to improve LoRA's performance by a discrimination-based prompting method, leveraging LLMs' In-Context Learning ability. \ours incorporates a matching network that selects prompts from a user-defined prompt pool, concatenates the selected prompts with the user instruction as input, and performs inference using the LoRA-fine-tuned LLMs. Trained with Reinforcement Learning, \ours exhibits commensurate performance on various evaluation metrics compared with typical instruction fine-tuning methods, utilizing only consumer-grade GPU resources and exhibiting a large reduction in computational costs. 

\end{abstract}


\input{sections/introduction}

\input{sections/method}

\input{sections/experiments}

\input{appendix/related_work}

\input{sections/conclusion}


\input{additions/limitations}

\input{additions/ethics_statement}


\bibliographystyle{acl_natbib}
\bibliography{main}

\clearpage
\appendix
\onecolumn

\input{appendix/examples}
\clearpage
\input{appendix/hyp}
\clearpage

\end{document}

%% file: sections/introduction.tex
\section{Introduction}

In recent years, the impressive achievements of large language models (LLMs) have become increasingly evident. Online LLM products, e.g., Claude \cite{bai2022training} and ChatGPT \cite{OpenAI2023GPT4TR}, have been widely recognized by the industry for their strong capabilities and are utilized in a myriad of industrial tasks \cite{liu2023summary, zhao2023survey}. The achievement of such success highly hinges on the usage of supervised fine-tuning (SFT) \cite{mishra2021cross, sanh2021multitask, wei2021finetuned}.

Nevertheless, as these models become larger, so does the intricacy of SFT. These fine-tuning procedures typically demand a large scale of computational resources to accommodate training all the model parameters. Consequently, this can be economically challenging for independent developers and smaller entities, who often have their own specific needs and budget limitations. In addition, data privacy standards prevent them from using third-party APIs, adding another layer of constraint for them to utilize the LLMs. Thus, while LLMs have been evolutionary in various applications, their scalability and cost-effectiveness still pose challenges in deployment.

To solve the aforementioned problem, some have applied parameter-efficient finetuning which updates a relatively small portion of parameters, making fine-tuning more manageable under resource limitation \cite{hu2021lora, dettmers2023qlora, glora, prefixtuning, prompttuning, ptuning}. \citet{hu2021lora} introduce LoRA to train dense layers by optimizing their rank decomposition matrices, thus considerably minimizing the number of trainable parameters and not adding to inference latency. However, LLM's performance may be limited as LoRA only trains a subset of the model parameters. Furthermore, LoRA may not achieve good performance on some tasks with unique characteristics because it can hardly adapt to diverse datasets due to its static
fine-tuning strategy \cite{glora}.

\begin{figure}[t] 
\includegraphics[width=0.495\textwidth]{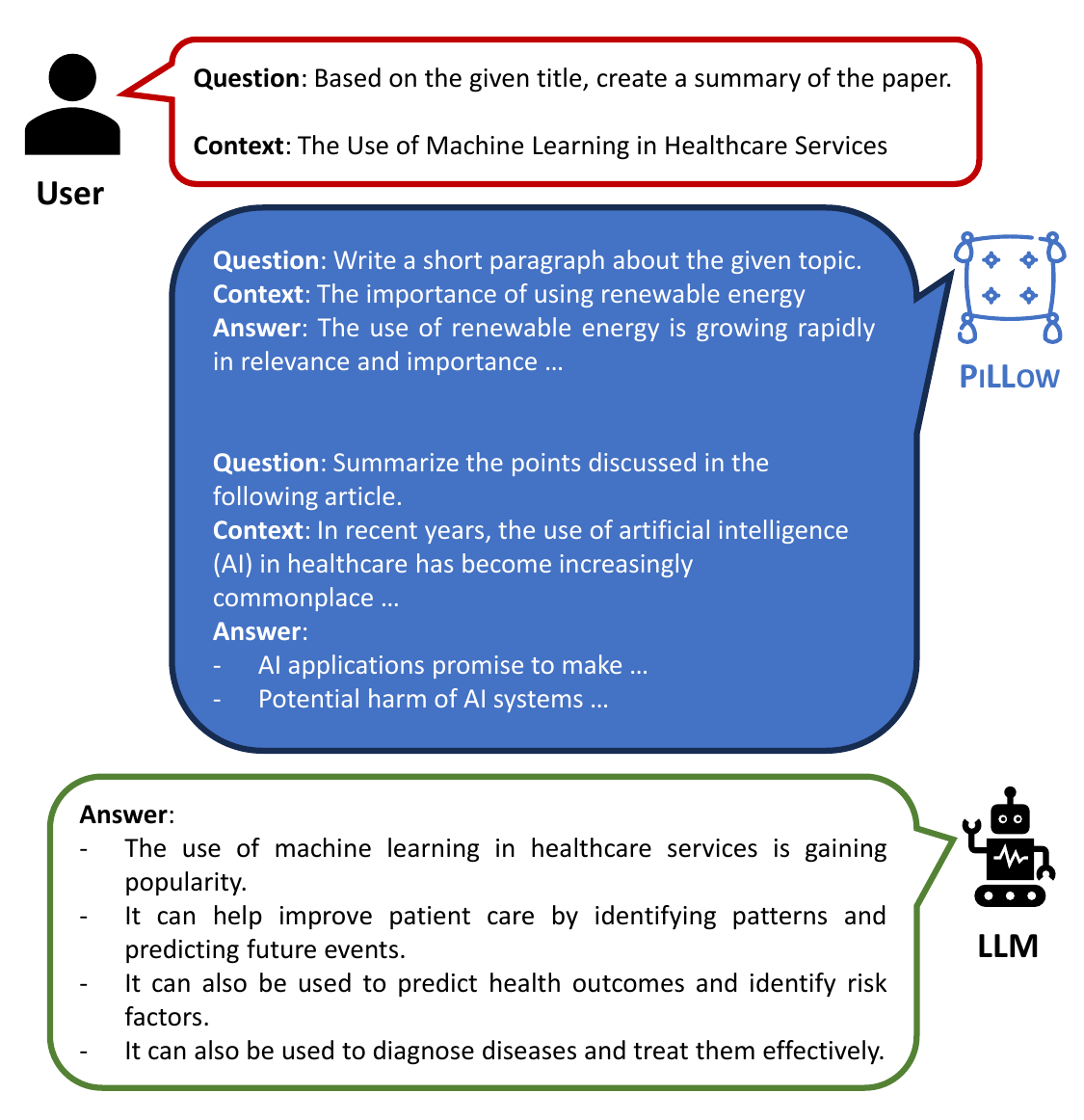}
\caption{A demonstration of 2-shot \ours.} 
\label{fig.demo} 
\end{figure}

Therefore, \textit{can we attain a similar performance level to SFT by merely employing a comparable amount of resources as used by LoRA? Can we realize it using LoRA-fine-tuned LLMs' reserved in-context learning (ICL) capacity?} Our approach, named \textbf{\ours}, trains a \textbf{P}rompt matching net using Re\textbf{I}nforcement \textbf{L}earning to improve fine-tuning LLMs with \textbf{LOW} computation resources. We train an RL agent to select exemplars from a comprehensive ``prompt set'' which can be defined by users or split from the training data, and subsequently merge these examples to form a prompt which is then added at the beginning of the input text that is fed into LoRA model. Figure \ref{fig.demo} shows an example where a 2-shot prompt is given by \ours agent based on the user input. Our approach becomes particularly beneficial in commercial applications where user input styles vastly deviate from those found in the pre-training corpus but are confined to limited variations. In these circumstances, the RL agent can efficiently learn to choose examples that best resonate with the specific query posed to the LLMs and therefore achieve comparable performance with direct SFT training. Our contributions can be summarized as follows:
\begin{itemize}
    \item We propose a new framework \ours to achieve SFT-comparable performance by utilizing LoRA and ICL with limited resources.
    \item We make \ours easy to use and widely applicable because a pre-trained LLM can be shared and used to build many LoRA adapters and matching networks for different tasks. 
    \item Experiments show that our proposed \ours is effective in instruction-finetuning datasets that contain diverse tasks in various domains.
\end{itemize}

%% file: sections/method.tex
\section{Method}
We present \ours, a novel RL-based prompting matching framework, designed to enhance the performance of the fine-tuned LoRA model leveraging in-context learning. To provide a better understanding of our work, we first give a brief overview of the necessary background information. Following that, we depict our task by framing it within the Reinforcement Learning settings and subsequently detail the components of our framework.

\subsection{Preliminary}
\subsubsection{Supervised Fine-tuning and LoRA}
The technique of supervised fine-tuning (SFT) is employed for enhancing the capabilities of pre-training language models by subjecting them to additional training on labeled datasets for the purpose of task-specific or domain-specific adaptation. This process involves the recalibration of model parameters by minimizing a defined loss function, thereby aligning its predictive capacity with the anticipated outputs. SFT takes advantage of accumulated prior knowledge to augment the efficiency in subsequent tasks, such as text categorization, named entity recognition, sentiment analysis, and etc. However, comprehensive fine-tuning LLMs through SFT becomes less practical as the size of LLMs increases, especially for individual developers and studios. A promising solution to this predicament is LoRA \cite{hu2021lora}, which proposes the training of rank decomposition matrices for each layer in the model architecture. This method significantly curtails the number of trainable parameters for subsequent tasks without imposing inference latency. However, for general instruction following tasks that experience large distributional shifts between different tasks, LoRA cannot achieve comparable performance with SFT due to the relatively low capacity. 

\subsubsection{In-Context Learning}
In-context learning (ICL) is a method that enhances LLMs by supplying specific contexts using a handful of examples, or prompts, to steer the model's behavior and produce the required results \cite{dong2022survey}. Efficient prompts can direct the model's responses by offering pertinent information via sentences, keywords, instructions, or examples. ICL enables users to tailor the model for specialized tasks or fields, leveraging a relatively smaller dataset composed of examples and intended outputs. Nevertheless, as \citet{zhao2021calibrate} emphasizes, ICL can be highly sensitive to the setup of prompts, encompassing prompt templates, in-context examples, and the order of the examples. (For more related work in prompting, please refer to Section \ref{sec:related_work})

\subsection{\ours}
\subsubsection{Motivation}
Our objective is to construct an interpretable and resource-efficient automated prompting framework. Despite the superior performance they exhibit, continuous prompting methods do not provide human interpretable results and mandate the utilization of costly gradient information \cite{promptsurvey}. Recent advancements in the discrete prompting field have brought forward generation-based \cite{deng2022rlprompt} and editing-based \cite{shin2020autoprompt, zhang2022tempera} methods which have demonstrated their efficacy across various task domains. However, these approaches encounter significant challenges in terms of their computational intensity during the training phase, which is the main issue \ours aims to address. 

On the premise that discrimination is much less computationally intensive than generation or editing, we propose to build a discrimination-based prompting framework. In essence, \ours aims to identify the optimal prompt that aligns with the user's input, as opposed to generating or editing one. To begin with, the process of training a matching neural network exhibits greater resource efficiency by eschewing the necessity for direct operation on texts. Secondly, many downstream tasks exhibit a restricted diversity of types of questions and answers, leading to a scenario where a multitude of user inputs can be guided with several analogous examples. For organizations operating under computational resource constraints, the establishment of a compact suite of ``standard'' question-answer pairs is sufficient to prompt LoRA fine-tuned LLM to accomplish a designated task via the ICL capacity reserved by LoRA.

\subsubsection{RL-based Prompt Matching}
\textbf{Prompt Matching Problem}~~~Our goal is to select a series of optimal prompts $V=\{v_1, ..., v_m\}$ from a user-defined prompt set $P = \{p_i\}_{i=0}^{n-1}$, where $m$ is the number of shots and $n$ is the size of the prompt set, to maximize some performance measure $R$. The $R$ should be defined domain-specifically and will be discussed in Section \ref{sec.exp.t2t.reward}. Each prompt $p_i$ to be selected is a triple of $(\rm question, \rm context, \rm answer)$, where $\rm   question$ represents the user instruction, $\rm context$ denotes the extra information provided by the users (optional), and $\rm answer$ is the expected output. We formulate the task of \textit{prompt matching} as follows: 
\begin{equation}
\underset{V \subset P}{\mathrm{max}}R(y_\mathrm{LM} \sim M_{\mathrm{LM}}(\cdot|v_0, v_1, ..., v_m,x)),
\end{equation}
where $v_0$ denotes a pre-defined initial system prompt and the response $y_\mathrm{LM}$ is sampled by the LoRA fine-tuned LLM $ M_{\mathrm{LM}}(\cdot|v_0, v_1, ..., v_m,x)$ given the condition of user instruction $x$ and the prompts $\{v_i\}_{i=0}^m$ added to its front.

\noindent
\textbf{RL Formulation}~~~The prompt matching task can be formulated as a Markov Decision Process (MDP) as follows: given an initial state $s_0=(v_0, x)$, at each time step $t$, an RL agent $\pi_\theta$ with parameter $\theta$ selects a prompt index $k=a_t$ from the action space $A$ according to policy $\pi_\theta(a_t|s_{<t}, x)$. We define the transition function as: $\mathcal{T}: S \times A \rightarrow S$ to be the state before and after selecting a new prompt $(v_0, ..., v_t) \times a_t \rightarrow (v_0, ..., v_t, v_{t+1})$, where $v_{t+1}=p_{k}$, and the process stops when $t=m$. Then, we can optimize the policy $\pi_\theta$ by maximize the cumulative rewards: 
\begin{equation}
\begin{aligned}
    \underset{\theta}{\mathrm{max}}&\mathbb{E}[\sum_{t=0}^m \gamma^t r(y_{{\mathrm{LM}},t})],{\rm s.t., }y_{{\mathrm{LM}},t} \sim M_{\mathrm{LM}}(\cdot|\hat{s_t},x),
\end{aligned}
\end{equation}
where $\hat{s_t}\sim \prod_{i=0}^{t} \pi_\theta(a_i|s_{<i}, x)$, $r$ is the reward measurement, and $\gamma$ is the discount factor. We discuss the necessity of using RL in our task in Section \ref{sec.ablation.rl}.

\noindent
\textbf{Action Space}~~~The action space is simply the set of the indices of all candidate prompts. We preprocess the prompt set $P$ by encoding its QA pairs into an embedding set $P^\prime=\{f_i\}_{i=0}^{n-1}$ so that each prompt index $k$ corresponds to one embedding vector $f_k$. Suppose we have $n$ user-defined candidate prompts, then at each stage, the agent chooses an integer from $0$ to $n-1$, and the discrete action space size will be $n$. 

\begin{figure*}[htp] 
\centering
\includegraphics[width=0.9\textwidth]{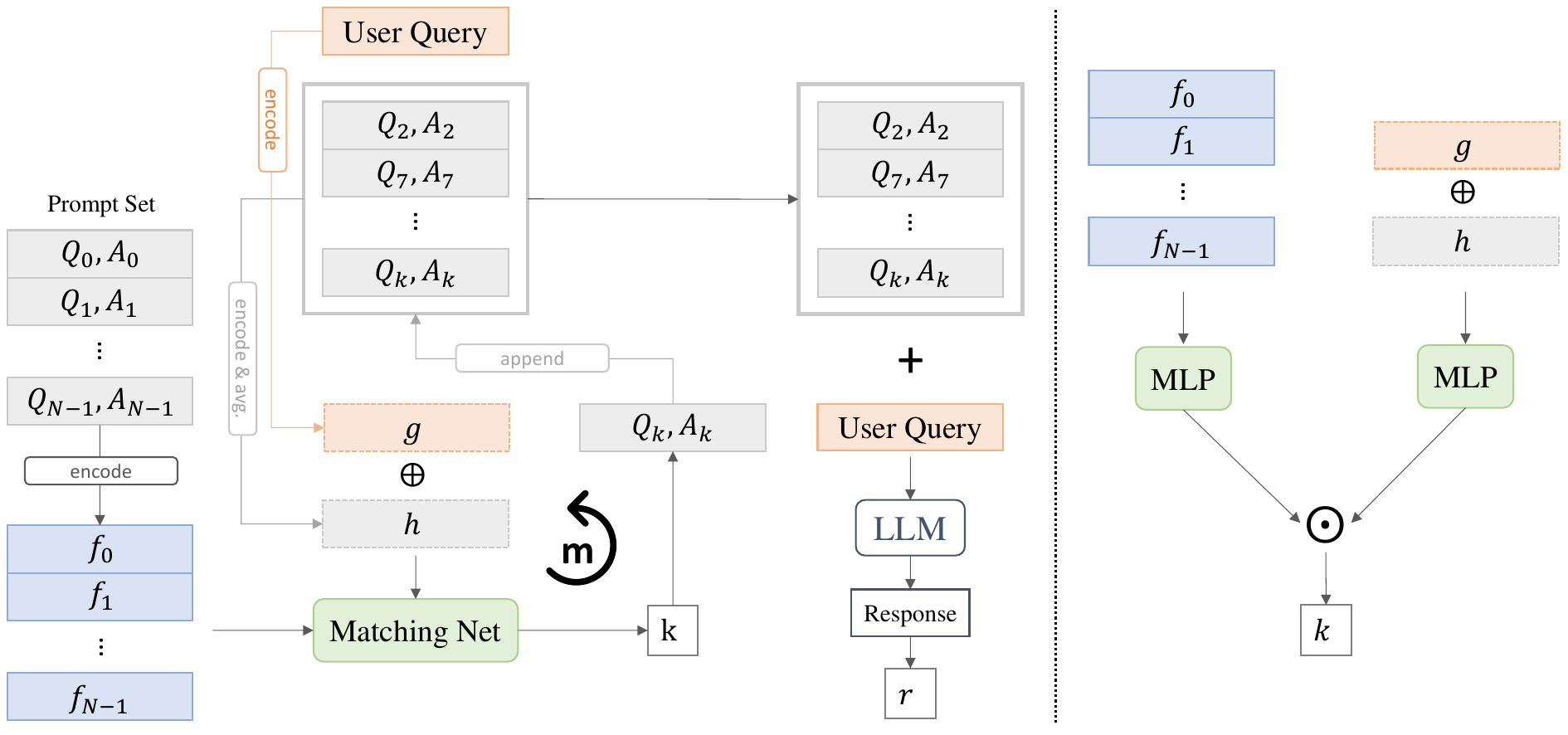}
\caption{Illustration of \ours. The left figure shows how the matching net is trained: At each step (out of $m$ steps), one prompt is selected from the prompt set by the matching net according to the user query and current matched prompts. After prompts are collected, they are passed to the LLM to get the answer, from which we calculate a reward. The right one shows the detailed pipeline of the matching network: The left MLP transforms the prompts into a set of vectors, with which we calculate dot products with the vector transformed by the right MLP from the state representation, and we obtain a probability distribution over the prompts. } 
\label{fig.prompter} 
\end{figure*}

\noindent
\textbf{State Representation}~~~Before matching, the user instruction input is encoded as an embedding $g$, and the embeddings of the selected prompts are aggregated and averaged to a new representation $h$. Then, we can get the state representation as $l=\mathrm{concat}(g, h)$ by concatenating two embeddings. To track state changes in the RL environment, we use a list of indices of chosen prompts instead of prompt context to further reduce the time and space complexity. The initial state will be a list containing $-1$ index only, i.e. $s_0 = [-1]$. As the episode proceeds, the list will be enlarged with new prompt indices appended, i.e. $s_t = \mathrm{append}(s_{t-1}, a_t)$. 

\noindent
\textbf{Policy Network}~~~We build the policy network $\pi_\theta(a_t|s_{<t}, x)$ with a deep text-matching network. The right-hand side of Figure \ref{fig.prompter} shows the network $Z_{\theta}$, which consists of two Multi-Layer Perceptrons (MLPs) to match two parts of features. The transformed prompt set $P^\prime=\{f_i\}_{i=0}^{n-1}$ is further encoded by an MLP: $c_i = Z_1(f_i)$, where $P^{\prime\prime}=\{c_i\}_{i=0}^{n-1}$ is named ``keys'', and the state representation $l$ is encoded by another MLP: $q = Z_2(l)$, where $q$ is named ``query''. We compute the similarities of the query and keys, scale by a normalization factor, and obtain a probability distribution $X$ after the softmax layer. Finally, we sample an integer number $k$ as the index of the matched prompt. 

\noindent
\textbf{Framework}~~~Based on the RL settings defined above, we design the entire training procedure as shown on the left-hand side of Figure \ref{fig.prompter}. Given $P^\prime$, $g$, and $h$, the RL agent $\pi_{\theta}$ consistently selects new prompt index $k$, looks up in the prompt set $P$, and appends the selected QA pair $p_k$ to previously selected prompts, until the number of prompts reaches $m$, which is the pre-defined number of shots. Then, all the $m$ selected prompts together with the user input are fed into the LoRA fine-tuned LLM $M_\mathrm{LM}$. The response will be scored by the reward function and the reward signal $r$ is used to update the parameters $\theta$ through off-the-shelf RL algorithms \cite{deng2022rlprompt, zhang2022tempera}. During the inference stage, the trained agent follows the same manner as aforementioned to select prompts and compose the LLM input.

%% file: sections/experiments.tex
\section{Experiments}
We conduct a comparative evaluation of our proposed framework \ours against two typical baseline methods: LoRA and SFT. SFT requires a high quantity of resources with high-quality response, while LoRA operates effectively under constrained resources but the performance is inferior to SFT. Nevertheless, empirical findings from our experimental studies suggest that \ours has the capability to yield performance in parity with SFT, even under low-resource constraints.

\input{tabs/whole}

\subsection{Datasets}
We use comprehensive instruction fine-tuning datasets that are designed to align the LLMs as helpful human assistants to follow almost all kinds of instructions. The following datasets are chosen because they encompass a variety of text-to-text generation tasks and contain repetitive QA patterns.

    \noindent
    \textbf{Alpaca} \cite{alpaca} contains 52,000 instructions and demonstrations which are generated by OpenAI's text-davinci-003 model given new prompts that explicitly outline the requirements, aiming at conducting instruction-tuning to make LLMs follow instructions better. Using Self-Instruct \cite{wang2022self}, the authors built the data generation pipeline to align pre-trained LMs with instructions generated by themselves. 
    
    \noindent
    \textbf{Dolly} \cite{DatabricksBlog2023DollyV2} is a human-annotated dataset of 15,000 instruction-following records, including various categories like brainstorming, classification, closed QA, generation, and summarization. The annotators are given instructions to refrain from using data from any online source except Wikipedia (for specific subsets of instruction categories), and most importantly, they avoid using generative AI in writing instructions or responses. 

\subsection{Reward Function}
\label{sec.exp.t2t.reward}
Since we test \ours on general text-to-text generation tasks, we simply use the weighted sum of textual similarity and semantic similarity as the score function $\zeta$ instead of conducting task-specific reward engineering \cite{deng2022rlprompt, zhang2022tempera}:
\begin{equation}
    \zeta(y, \hat{y}) = \lambda \cdot S_{\mathrm{textual}}(y, \hat{y}) + (1-\lambda) \cdot S_{\mathrm{semantic}}(y, \hat{y}),
\end{equation}
where $S_{\mathrm{textual}}$, $S_{\mathrm{semantic}}$ are textual similarity (based on fuzzy matching) and semantic similarity (based on cosine similarity between sentence representations), and $y$, $\hat{y}$ are LLM's output and expected output, respectively, and $\lambda$ is a balancing factor. Note that in deployment, people can choose desired reward formulations based on their specific tasks. 
Finally, the reward $r$ is obtained by scaling the score with a constant $\alpha$:
$r = \alpha \cdot \zeta(y, \hat{y})$.

\subsection{Experiment Setup}
We use Bloomz-560m, Bloomz-1b1, and Bloomz-7b1 \cite{muennighoff2022crosslingual} as backbone models to show \ours's effectiveness on LLMs of different sizes. For both Alpaca and Dolly, we use the entire dataset for LoRA/SFT training. Then, we randomly select 1,200 data items: 100 for the user-defined prompt set, 900 for RL training, and 200 for testing. For model training, we implement the LoRA-/\ours-related experiments on one V100 GPU and SFT with one A100 GPU for efficiency. We conduct the evaluation experiments of \ours on one NVIDIA GeForce RTX 3090 GPU.

\subsection{Evaluation}
\textbf{Automatic Scores}~~~We automatically score the LLM output by reward (r/w), ChatGPT score (C-Score), and perplexity (PPL). We eliminate abnormal values and then take the average to obtain the metric measurement. For C-Score evaluation, we utilize the prompt introduced by \citet{zhou2023lima} and reduce the score bias by randomly organizing the instruction and response orders \cite{wang2023large}.

\noindent
\textbf{Manual Scores}~~~We also evaluate the LLM output with a manual score (M-Score). This process is an absolute analysis which is similar to the method used by \citet{zhou2023lima}. We invite five human annotators to label each response with three labels: \textbf{Excellent}, \textbf{Pass}, \textbf{Fail}, which have the same criteria as \citet{zhou2023lima}. For each experiment, we randomly select 50 samples for labeling.

\subsection{Results}
We present our experiment results on 1-shot \ours in Table \ref{tab:whole}. We report the M-Score by reporting the average rate from human annotators in the order of \textbf{Excellent}/\textbf{Pass}/\textbf{Fail}. It can be seen that our method outperforms the LoRA model on both Alpaca and Dolly across most evaluation metrics and achieves performance very close to SFT. 

We observe that as the model size increases, the performance gain compared with LoRA tends to become larger. On Alpaca, for example, the 1b1 model trained with \ours surpasses LoRA by 0.12 in C-Score and 11.45 in perplexity, and for the 7b1 model such gaps increase to 0.49 and 17.05, respectively. 
Also, we can see that \ours helps large models reach very close performance to SFT. On Dolly, for example, the 1b1 model trained with \ours reaches 98.22\% of SFT's performance in ChatGPT score and 90.46\% in reward, and the 7b1 model reaches 97.94\% and 90.55\%, respectively. The human evaluation results also demonstrate similar pattern with the C-Score evaluation. 

Our observations in M-Score indicate that \ours significantly enhances the \textbf{Excellent} rate while reducing the \textbf{Fail} rate when compared to the LoRA model. This signifies a considerable improvement in quality in comparison to the LoRA model. Note that the 560m model trained with both \ours and LoRA does not improve that much and even degrades a little bit compared with the original pre-trained model, and on Alpaca \ours even performs worse than LoRA on the reward metric. However, there is no such problem for the 1b1 and 7b1 models. Therefore, we can conclude that our proposed \ours is particularly suitable for large-scale LLMs which inherently possess enhanced ICL and emergent capabilities. Importantly, the application of LoRA does not diminish these intrinsic abilities of the LLMs. We refer readers to Appendix \ref{sec:eg} for example LLM inputs and outputs.

\vspace{-1mm}
\subsection{Ablation Studies}
\subsubsection{Why RL?}
\label{sec.ablation.rl}
RL techniques have been widely used in multiple industrial application and achieved significant improvement in numerous domain \cite{qu2023bellman,coronato2020reinforcement,qiu2022latent,chen2022model,hambly2023recent}. Some may wonder why RL is even necessary in our settings since it seems that the tasks can be solved by simply matching the prompt with the largest similarity with the question. In this section, we conduct ablation study to show the superiority of RL in our setting.

As can be seen in Table \ref{tab.ab.rl}, results reveal that \ours outperforms simple matching (SimMatch) and LoRA model. This is due to the potential misalignment between the sentence encoder and LLM training data---the most semantically matched prompt might not yield the best answer. The sentence encoder and the LLM may not be trained on the same data distribution, so the prompt chosen by the encoder that best matches the question, i.e. seemed ``matched'' to humans, may not seem that ``matched'' to the LLM. Even if the chosen prompt is truly the most semantically similar one to the question, it may not best prompt to assist the LLM in generating an appropriate answer. In addition, users might want to switch to different LLMs to adapt to different downstream tasks, where such ``appropriateness" might be defined differently, therefore an RL-based training framework is necessary.

\input{tabs/ablation_rl}


\subsubsection{Number of Shots}

We also investigate the impact of increasing the number of ``shots''. 
We ablate on the number of shots to study how the number of exemplars affects the performance. As can be seen in Table \ref{tab.ab.shots}, as we increase the number of shots from 1 to 3, the ChatGPT score and reward increase a little, but in general, the measurements do not change too much. Intuitively, the more exemplars are given, the better the LLM output would be. However, it imposes difficulty to the matching net to ensure the selected exemplars are all helpful for prompting the LLM. It is possible that an irrelevant prompt is newly chosen and corrupts the LLM output. Also, a large number of shots makes the RL training slow. Therefore, we recommend using a small number of number of shots to balance the negative effects. In practice, we recommend just implement one prompt for \ours with best efficiency. 

\input{tabs/ablation_shots}


%% file: tabs/whole.tex
\begin{table*}[t]
\centering
\begin{tabular}{lllllll}
\toprule[2pt]
\textbf{Dataset}         & \textbf{Model Size}               & \textbf{Method} & \textbf{C-Score} & \textbf{PPL} & \textbf{R/w} & \textbf{M-Score} \\ \midrule[1pt]
\multirow{12}{*}{Alpaca} & \multirow{4}{*}{560m} & -               & 2.71                   & 192.24              & 4.87            &          0.00/0.16/0.84             \\
                         &                              & SFT             & 2.87                   & 106.07              & 5.30            &   0.04/0.40/0.56                    \\
                         &                              & LoRA            & 2.56                   & 149.93              & 4.89            &    0.00/0.32/0.68                   \\ \cline{3-7} 
                         &                              & \textbf{\ours}            & 2.63 \textcolor{red}{(+0.07)}           & 140.57 \textcolor{red}{(-9.36)}      & 4.68 \textcolor{blue}{(-0.21)}    &       0.02/0.21/0.77                \\ \cline{2-7} 
                         & \multirow{4}{*}{1b1}  & -               & 3.01                   & 108.2               & 5.71            &         0.00/0.17/0.83              \\
                         &                              & SFT             & 3.29                   & 52.02               & 6.48            &           0.12/0.43/0.45            \\
                         &                              & LoRA            & 3.09                   & 78.81               & 5.83            &              0.09/0.21/0.70           \\ \cline{3-7} 
                         &                              & \textbf{\ours}            & 3.21 \textcolor{red}{(+0.12)}           & 67.36 \textcolor{red}{(-11.45)}      & 5.89 \textcolor{red}{(+0.06)}    &     0.14/0.39/0.47                  \\ \cline{2-7} 
                         & \multirow{4}{*}{7b1}  & -               & 3.14                   & 161.19              & 5.88            &      0.00/0.23/0.77                 \\
                         &                              & SFT             & 3.84                   & 64.34               & 6.49            &            0.31/0.54/0.15           \\
                         &                              & LoRA            & 3.27                   & 120.70              & 5.94            &         0.17/0.46/0.37              \\ \cline{3-7} 
                         &                              & \textbf{\ours}            & 3.76 \textcolor{red}{(+0.49)}           & 103.65 \textcolor{red}{(-17.05)}     & 6.07 \textcolor{red}{(+0.13)}    &                 0.29/0.44/0.27      \\ \midrule[1pt]
\multirow{12}{*}{Dolly}  & \multirow{4}{*}{560m} & -               & 2.83                   & 247.45              & 4.26            &           0.00/0.18/0.82            \\
                         &                              & SFT             & 3.01                   & 218.61              & 5.01            &             0.07/0.42/0.51          \\
                         &                              & LoRA            & 2.64                   & 221.16              & 4.34            &              0.00/0.33/0.67         \\ \cline{3-7} 
                         &                              & \textbf{\ours}            & 2.74 \textcolor{red}{(+0.1)}            & 191.9 \textcolor{red}{(-29.26)}      & 4.70 \textcolor{red}{(+0.36)}    &     0.05/0.39/0.56                  \\ \cline{2-7} 
                         & \multirow{4}{*}{1b1}  & -               & 3.13                   & 227.43              & 4.74            &              0.00/0.19/0.81         \\
                         &                              & SFT             & 3.37                   & 67.93               & 5.87            &           0.14/0.51/0.35            \\
                         &                              & LoRA            & 3.08                   & 140.40              & 4.79            &           0.07/0.32/0.61            \\ \cline{3-7} 
                         &                              & \textbf{\ours}            & 3.31 \textcolor{red}{(+0.23)}           & 112.78 \textcolor{red}{(-27.62)}     & 5.31 \textcolor{red}{(+0.52)}    &                 0.11/0.47/0.42      \\ \cline{2-7} 
                         & \multirow{4}{*}{7b1}  & -               & 3.24                   & 244.09              & 4.86            &               0.00/0.26/0.73        \\
                         &                              & SFT             & 3.89                   & 56.64               & 5.61            &              0.39/0.51/0.10         \\
                         &                              & LoRA            & 3.33                   & 146.92              & 4.93            &              0.21/0.48/0.31         \\ \cline{3-7} 
                         &                              & \textbf{\ours}            & 3.81 \textcolor{red}{(+0.48)}           & 113.09 \textcolor{red}{(-33.83)}     & 5.08 \textcolor{red}{(+0.15)}    &     0.36/0.47/0.17                  \\ \bottomrule[2pt]
\end{tabular}
\caption{Results on 1-shot \ours on Alpaca and Dolly. The score differences that indicate better performance than LoRA are marked with \textcolor{red}{red} color, while those showing worse performance are marked with \textcolor{blue}{blue} color.}
\label{tab:whole}
\end{table*}

%% file: tabs/ablation_rl.tex
\begin{table}[H]
\small
\centering
\begin{tabular}{lllll}
\toprule[2pt]
\textbf{Method} & \textbf{C-Score} & \textbf{PPL} & \textbf{R/w} & \textbf{M-Score} \\ \midrule[1pt]
LoRA            & 3.09             & 78.81        & 5.83            &           0.09/0.21/0.70         \\
SimMatch        & 3.12             & 79.26        & 5.81            &       0.12/0.26/0.62           \\
\ours            & \textbf{3.21}             & \textbf{67.36}        & \textbf{5.89}            &          \textbf{0.14}/\textbf{0.39}/\textbf{0.47}      \\ \bottomrule[2pt]
\end{tabular}
\caption{Ablation on prompting framework. Experiments are conducted with Bloomz-1b1 on Alpaca test set.}
\label{tab.ab.rl}
\end{table}

%% file: tabs/ablation_shots.tex
\begin{table}[H]
\small
\centering
\begin{tabular}{lllll}
\toprule[2pt]
\textbf{Type} & \textbf{C-Score} & \textbf{PPL} & \textbf{R/w} & \textbf{M-Score} \\ \midrule[1pt]
1-shot           & 3.21             & 67.36        & 5.89            &        0.14/0.39/0.47          \\
2-shot           & 3.19             & \textbf{66.95}        & 6.05            &      0.14/\textbf{0.41}/\textbf{0.45}            \\
3-shot           & \textbf{3.23}             & 69.92        & \textbf{6.17}            &      \textbf{0.16}/0.38/0.46            \\ \bottomrule[2pt]
\end{tabular}
\caption{Ablation on the number of shots. Experiments are conducted with Bloomz-1b1 on Alpaca test set.}
\label{tab.ab.shots}
\end{table}

%% file: appendix/related_work.tex
\section{Related Work}
\label{sec:related_work}

Since manually writing prompts is time-consuming and labor-intensive, a number of methods have been proposed to automate the prompting process. In continuous prompting (a.k.a soft prompting) \cite{promptsurvey}, prompting is performed directly in the embedding space of the language models. However, by their continuous nature, such prompts are not human-understandable. \textit{Prefix Tuning} \cite{prefixtuning} adds a sequence of continuous task-specific prompt embeddings to the front of input texts in each layer of LM while keeping the LM's parameters frozen. Similarly, \textit{Prompt Tuning} \cite{prompttuning} prepends the input texts with special tokens to form a template and directly tune the token embeddings without updating LM's parameters. Unlike the two methods, \textit{P-Tuning} \cite{ptuning} removes the restriction on adding the prompt embedding to the beginning of the input. They define that the prompt tokens can be inserted anywhere in the input sequence and can only be inserted in the input rather than any other model layer. 

Approaches on discrete prompting (a.k.a hard prompting) \cite{promptsurvey} automatically generate or edit prompts described in a discrete space, i.e. in the form of texts. \textit{AutoPrompt} \cite{shin2020autoprompt} edits textual prompt template in a gradient-guided manner, and find that the best final prompts are usually gibberish and not human-interpretable. \textit{TEMPERA} \cite{zhang2022tempera} is also an editing-based method, but it trains the test-time editor with RL framework and edits the initial prompts using commonly-used instructions, few-shot exemplars, and verbalizers. Similarly, \textit{RLPrompt} \cite{deng2022rlprompt} is also built on an RL framework, which generates better prompts word by word with black-box optimization. The authors also find that final optimal prompts are often ungrammatical texts and they are transferrable between different LMs. However, both generation and editing are hard tasks and can be computationally intensive given their large action space and long decision process. Also, the RL-based methods rely on specific reward designs, which only apply to limited tasks like few-shot text classification.

Recent work has also leveraged pre-defined exemplar pools to boost the final performance of prompting LLMs. \citet{rubin2021learning} trained a dense retriever that fetches useful training examples as LLM prompts from an exemplar pool during test time. \citet{liu2021makes} suggest retrieving training pool exemplars that are semantically comparable to a test example, and they demonstrate how this can greatly improve performance. Similarly, \textit{TEMPERA} \cite{zhang2022tempera} design an attention-based exemplar selector over the embedding space and show that such an exemplar selection process can effectively choose training examples that lead to high performance.

%% file: sections/conclusion.tex
\section{Conclusion}

We train a prompt matching framework \ours via Reinforcement Learning to enhance efficient instruction finetuning. \ours is evaluated on the most recent instruction finetuning datasets, Alpaca and Dolly, and achieves superior results across all evaluation metrics and model sizes compared with supervised fine-tuning under LoRA. This new area of research combining prompting, matching, and RL can inspire future work on better prompting methods for LLMs under low-resource regimes.

%% file: additions/limitations.tex
\section*{Limitations}
\ours is implemented based on matching rather than generation. Despite being highly controllable, interpretable, and efficient, such a prompting method may not show superior performance on large but sparse datasets in which most question-answer pairs do not follow similar patterns because the RL agent may not be able to find appropriate prompts. In our future work, we intend to investigate the impact on \ours's performance by utilizing a variety of reinforcement learning (RL) algorithms. Furthermore, we are interested in exploring hybrid RL agents with the aim to optimize the number of shots and prompts for each slot.

In terms of the reward design, we only include a semantic similarity and a textual similarity, which may not be enough for giving authentic feedback to the RL agent. We believe that other popular automatic text generation evaluation metrics such as BLEURT~\cite{sellam-etal-2020-bleurt}, BARTScore~\cite{yuan2021bartscore}, and GPTScore~\cite{fu2023gptscore} can also be utilized for such purpose.

%% file: additions/ethics_statement.tex
\section*{Ethics Statement}

We declare that the current study strictly comply with the \href{https://www.aclweb.org/portal/content/acl-code-ethics}{ACL Ethics Policy}. The datasets \cite{alpaca, DatabricksBlog2023DollyV2} used to compare \ours with previous methods are publicly available and we did not modify any data in them. For the manual evaluation, we anonymously hire 5 experts. We make scoring each LLM output as a unit task and pay \$0.2 for each unit task. On average, one human evaluator can finish 30 unit tasks per hour after short training and practice. We recommend that human evaluators devote a maximum of 2 hours per day to the evaluation work in order to maintain a comfortable pace.

%% file: appendix/examples.tex
\section{Example LLM Input \& Output}
\label{sec:eg}

We provide example LLM input and output here. The results are generated by Bloomz-7b1. To fit the page, we only showcase 1-shot prompts.

\begin{sidewaystable}[ht]
\tiny
\centering
\begin{tabular}{llll}
\toprule[2pt]
\textbf{User Input}                                                                                            & \textbf{Matched Prompts}                                                                                                                                                                                                                                                                                                                                                                                                                                                                                                    & \textbf{LLM Input}                                                                                                                                                                                                                                                                                                                                                                                                                                                                                                                                                                           & \textbf{LLM Output}                                                                                                                                                                                                                                                                                                                                                                                                                                                                       \\ \midrule[1pt]
What sport is Ruud Gullit known for?                                                                           & \begin{tabular}[c]{@{}l@{}}What is a polygon?\\ \\ \#\#\#\\ \\ A polygon is a form in Geometry.  \\ It is a single dimensional plane made of connecting lines \\ and any number of vertices.  \\ It is a closed chain of connected line segments or edges.  \\ The vertices of the polygon are formed where two edges meet.  \\ Examples of polygons are hexagons, pentagons, and octagons.  \\ Any plane that does not contain edges or vertices is not a polygon.\\ An example of a non-polygon is a circle.\end{tabular} & \begin{tabular}[c]{@{}l@{}}What is a polygon?\\ \\ \#\#\#\\ \\ A polygon is a form in Geometry.  \\ It is a single dimensional plane made of connecting lines \\ and any number of vertices.  \\ It is a closed chain of connected line segments or edges.  \\ The vertices of the polygon are formed where two edges meet.  \\ Examples of polygons are hexagons, pentagons, and octagons.  \\ Any plane that does not contain edges or vertices is not a polygon.\\ An example of a non-polygon is a circle.\\ \\ \#\#\#\\ \\ What sport is Ruud Gullit known for\\ \\ \#\#\#\end{tabular} & \begin{tabular}[c]{@{}l@{}}Ruud Gullit is a football player \\ who played for Barcelona and England.\end{tabular}                                                                                                                                                                                                                                                                                                                                                                         \\ \hline
\begin{tabular}[c]{@{}l@{}}Rewrite the sentence using an idiom: \\ The light suddenly disappeared\end{tabular} & \begin{tabular}[c]{@{}l@{}}Evaluate this sentence for spelling and grammar mistakes\\ He finnished his meal and left the resturant\\ \\ \#\#\#\\ \\ He finished his meal and left the restaurant.\end{tabular}                                                                                                                                                                                                                                                                                                              & \begin{tabular}[c]{@{}l@{}}Evaluate this sentence for spelling and grammar mistakes\\ He finnished his meal and left the resturant\\ \\ \#\#\#\\ \\ He finished his meal and left the restaurant.\\ \\ \#\#\#\\ \\ Rewrite the sentence using an idiom\\ The light suddenly disappeared\\ \\ \#\#\#\end{tabular}                                                                                                                                                                                                                                                                             & The light suddenly went out.                                                                                                                                                                                                                                                                                                                                                                                                                                                              \\ \hline
\begin{tabular}[c]{@{}l@{}}Explain how design thinking helps \\ organizations to innovate.\end{tabular}        & \begin{tabular}[c]{@{}l@{}}Describe the structure of an atom.\\ \\ \#\#\#\\ \\ An atom is made up of a nucleus, \\ which contains protons and neutrons, \\ surrounded by electrons that travel in orbits \\ around the nucleus. The protons and neutrons \\ have a positive charge, while the electrons \\ have a negative charge, resulting in an overall \\ neutral atom. The number of each particle \\ determines the atomic number and the type of atom.\end{tabular}                                                  & \begin{tabular}[c]{@{}l@{}}Describe the structure of an atom.\\ \\ \#\#\#\\ \\ An atom is made up of a nucleus, \\ which contains protons and neutrons, \\ surrounded by electrons that travel in orbits \\ around the nucleus. The protons and neutrons \\ have a positive charge, while the electrons \\ have a negative charge, resulting in an overall \\ neutral atom. The number of each particle \\ determines the atomic number and the type of atom.\\ \\ \#\#\#\\ \\ Explain how design thinking helps organizations to innovate.\\ \\ \#\#\#\end{tabular}                         & \begin{tabular}[c]{@{}l@{}}Design thinking is a process \\ used to solve problems through \\ creative problem solving. It involves \\ gathering information about the problem, \\ brainstorming solutions, testing \\ different ideas, evaluating them, \\ and making decisions based on the \\ results. Design thinking can help \\ organizations to innovate by helping \\ them to identify new opportunities \\ and develop innovative solutions for \\ their challenges.\end{tabular} \\ \bottomrule[2pt]
\end{tabular}
\end{sidewaystable}

%% file: appendix/hyp.tex
\section{Hyperparameters}
\label{sec:hyp}

We set the following hyperparameters for \ours evaluation:

\begin{table*}[h]
\large
\centering
\begin{tabular}{ll}
\toprule[2pt]
\textbf{Field}                 & \textbf{Value} \\ \midrule[1pt]
LoRA rank                      & 8              \\
number of RL traning epochs    & 150            \\
MLP input sizes                 & 384, 768        \\
MLP hidden size                & 1024           \\
MLP output size                & 512            \\
learning rate                  & 1e-6           \\
trainig batch size             & 32             \\
lambda (balancing factor)      & 0.2            \\
LLM number of beams            & 1              \\
LLM top p                      & 0.8            \\
LLM top k                      & 0              \\
LLM do sample                  & False          \\
LLM number of return sequences & 1              \\
LLM temperature                & 1              \\
LLM repetition penalty         & 1              \\
LLM max new tokens             & 512            \\
LLM length penalty             & 1              \\
LLM early stopping             & True           \\ \bottomrule[2pt]
\end{tabular}
\end{table*}